\documentclass[runningheads]{llncs}
\usepackage[T1]{fontenc}
\usepackage{url}
\usepackage{graphicx}
\usepackage{multirow}
\usepackage{amsmath}
\usepackage{booktabs,subcaption,amsfonts,dcolumn}
\newcolumntype{d}[1]{D..{#1}}
\usepackage{xcolor,colortbl}
\usepackage[misc,geometry]{ifsym} 
\definecolor{Gray}{gray}{0.85}

\newcolumntype{a}{>{\columncolor{Gray}}c}
\begin{document}
\title{MedSegDiff: Medical Image Segmentation with Diffusion Probabilistic Model}
\author{Junde Wu\inst{1} \and Rao Fu \inst{2}\and
Huihui Fang\inst{1} \and
Yu Zhang\inst{2} \and Yehui Yang\inst{1} \and Haoyi Xiong\inst{1} \and Huiying Liu\inst{3} \and Yanwu Xu\inst{1} \textsuperscript{ (\Letter)}}

\authorrunning{J. Wu et al.}

\institute{Baidu Research
\\
\and
Mind Vogue Lab\\
\and
Institute for Infocomm Research, A*STAR\\
}
\maketitle              
\begin{abstract}
Diffusion probabilistic model (DPM) recently becomes one of the hottest topic in computer vision. Its image generation application such as Imagen, Latent Diffusion Models and Stable Diffusion have shown impressive generation capabilities, which aroused extensive discussion in the community. Many recent studies also found it is useful in many other vision tasks, like image deblurring, super-resolution and anomaly detection. Inspired by the success of DPM, we propose the first DPM based model toward general medical image segmentation tasks, which we named MedSegDiff. In order to enhance the step-wise regional attention in DPM for the medical image segmentation, we propose dynamic conditional encoding, which establishes the state-adaptive conditions for each sampling step. We further propose Feature Frequency Parser (FF-Parser), to eliminate the negative effect of high-frequency noise component in this process. We verify MedSegDiff on three medical segmentation tasks with different image modalities, which are optic cup segmentation over fundus images, brain tumor segmentation over MRI images and thyroid nodule segmentation over ultrasound images. The experimental results show that MedSegDiff outperforms state-of-the-art (SOTA) methods with considerable performance gap, indicating the generalization and effectiveness of the proposed model. Our code is released at \url{https://github.com/WuJunde/MedSegDiff}.
\end{abstract}


\section{Introduction}
Medical image segmentation is the process of partitioning a medical image into meaningful regions. Segmentation is a fundamental step in many medical image analysis applications such as diagnosis, surgical planning, and image-guided surgery. This is important because it allows doctors and other medical professionals to better understand what they're looking at. It also makes it easier to compare images and track changes over time. In recent years, there has been a growing interest in automatic medical image segmentation methods. These methods have the potential to reduce the time and effort required for manual segmentation, and to improve the consistency and accuracy of results. With the development of the deep learning techniques, more and more studies successfully applied the neural network (NN) based models to the medical image segmentation tasks, from the popular convolution neural networks (CNN) \cite{ji2021learning} to the recent vision transformers (ViT) \cite{chen2021transunet,wang2021transbts,liu2022remote,zhao2021high}.


Very recently, diffusion probabilistic model (DPM)\cite{ho2020denoising} gained popularity as a powerful class of generative models\cite{zhao2021text}, that is able to generate images with high diversity and synthesis quality. Recent large diffusion models, such as DALL-E2\cite{ramesh2022hierarchical}, Imagen\cite{saharia2022photorealistic}, and Stable Diffusion\cite{rombach2022high} have shown incredible generation capability. 
Diffusion models are originally applied in fields in which there is no absolute ground-truth. However, recent studies show that it is also effective for the problems in which the ground-truth is unique, like super-resolution\cite{saharia2022image} and deblurring\cite{whang2022deblurring}. 

Inspired by the recent success of DPM, we design a unique DPM-based segmentation model for the medical image segmentation tasks. To our knowledge, we are the first to propose the DPM-based model under the background of general medical image segmentation with different image modalities. We note that in tasks of medical image segmentation, the lesions/organs are often ambiguous and hard to discriminate from the background. In that case, an adaptive calibration process is the key to obtain a delicate result. Following this mindset, we propose dynamic conditional encoding over vanilla DPM to design the proposed model, named MedSegDiff. Note that in the iterative sampling process, MedSegDiff conditions each of the step with image prior, in order to learn the segmentation map from it. Toward the adaptive regional attention, we integrate the segmentation map of current step into the image prior encoding at each step. The specific implementation is to fuse the current-step segmentation mask with the image prior on the feature level with a multi-scale manner. In this way, the corrupted current-step mask helps to dynamically enhance the condition features, thus improves the reconstruction accuracy. In order to eliminate the high-frequency noises in the corrupted given mask in this process, we further propose the feature frequency parser (FF-Parser) to filter the features in the Fourier space. FF-Parsers are adopted on each skip connection path for the multi-scale integration. We verify MedSegDiff on three different medical segmentation tasks, the optic-cup segmentation, the brain tumor segmentation, and the thyroid nodule segmentation. The images of these tasks have different modalities, which are the fundus images, brain CT images, the ultrasound images respectively. MedSegDiff outperforms the previous SOTA on all three tasks with different modalities, which shows the generalization and effectiveness the proposed method. In brief, the contributions of the paper are:

\begin{itemize}
\item The fist to propose DPM-based model toward general medical image segmentation.

\item Dynamic conditional encoding strategy is proposed for step-wise attention.

\item FF-Parser is proposed to eliminate the negative effects of high-frequency components.

\item SOTA performance on three different medical segmentation tasks with different image modalities. 
\end{itemize}

\section{Method}
We design our model based on diffusion model mentioned in\cite{ho2020denoising}. Diffusion models are generative models composed of two stages, a forward diffusion stage and a reverse diffusion stage. In the forward process, the segmentation label $x_{0}$ is gradually added Gaussian noise through a series of steps $T$. In the reverse process, a neural network is trained to recover the original data by reversing the noising process, which can be represented as:
\begin{equation}
    p_{\theta}(x_{0:T-1}|x_{T}) = \Pi^{T}_{t=1} p_{\theta}(x_{t-1}|x_{t}),
\end{equation}
where $\theta$ is reverse process parameters. Starting from a Gaussian noise, $p_{\theta}(x_{T}) = \mathcal{N}(x_{T};0,I_{n \times n})$, where $I$ is the raw image, the reverse process transforms the latent variable distribution $p_{\theta}(x_{T})$ to the data distribution $p_{\theta}(x_{0})$. To be symmetrical to the forward process, the reverse process recovers the noise image step by step to obtain the final clear segmentation.

\begin{figure*}
    \centering
    \includegraphics[width=\linewidth]{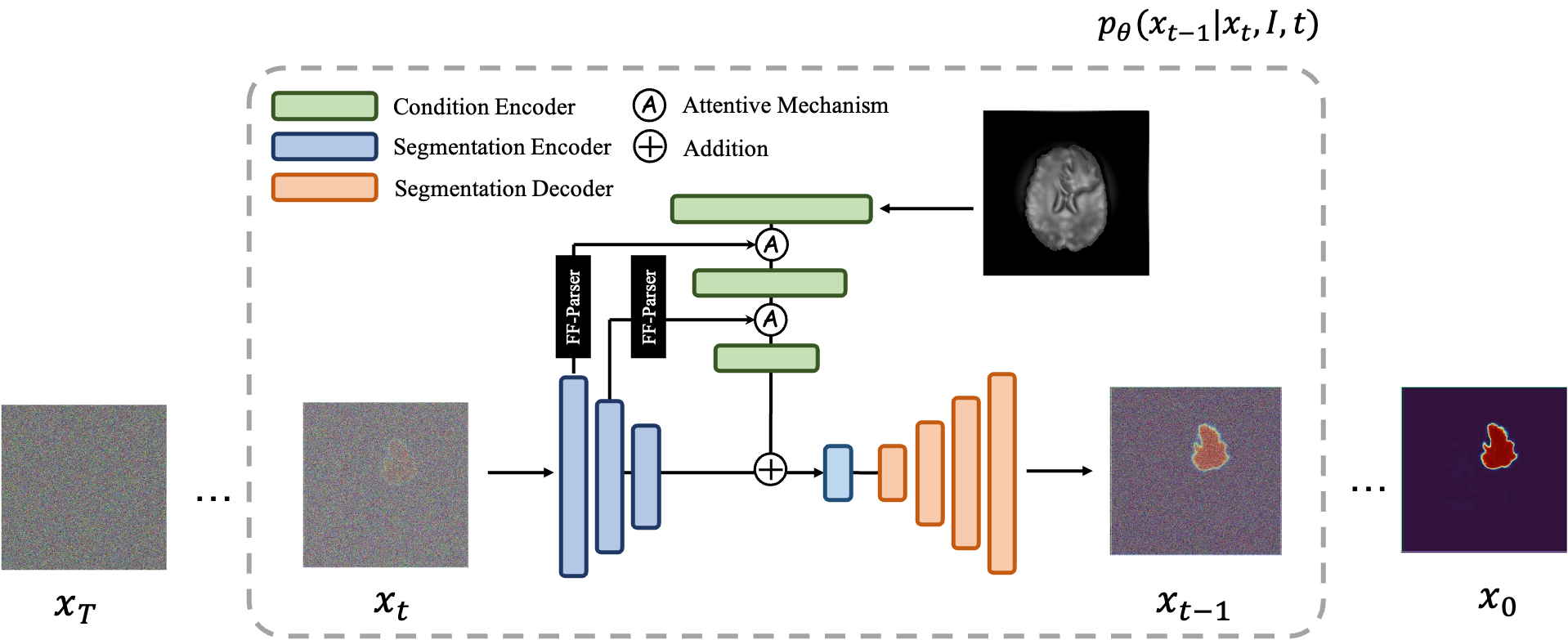}
    \caption{An illustration of MedSegDiff. For the clarity, the time step encoding is omitted in the figure.}
    \label{fig:framework}
\end{figure*}

Following the standard implementation of DPM, we adopt a UNet as the network for the learning. An illustration is shown in Figure \ref{fig:framework}. In order to achieve the segmentation, we condition the step estimation function $\epsilon$ by raw image prior, which can be represented as:
\begin{equation}
    \epsilon_{\theta}(x_{t},I,t) = D((E_{t}^{I} + E_{t}^{x},t),t),
\end{equation}
where $E_{t}^{I}$ is the conditional feature embedding, in our case, the raw image embedding, $E_{t}^{x}$ is the segmentation map feature embedding of the current step. The two components are added and sent to a UNet decoder $D$ for the reconstruction. The step index $t$ is integrated with the added embedding and decoder features. In each of these, it is embedded using a shared learned look-up table, following \cite{ho2020denoising}.

\subsection{Dynamic Conditional Encoding}
In most conditional DPM, the conditional prior will be a unique given information. However, medical image segmentation is notorious for its ambiguous objects. The lesions or tissues are often hard to discriminate from its background. The low-contrast image modalities, such as MRI or ultrasound images, make it even worse. Given only a static image $I$ as the condition for each step will be hard to learn. To address this problem, we propose a dynamic conditional encoding for each step. We note that on the one hand, the raw image contains the accurate segmentation target information but hard to discriminate from the background, on the other hand, the current-step segmentation map contains the enhanced target regions but not accurate. This motivated us to integrate the current-step segmentation information $x_{t}$ into the conditional raw image encoding for the mutual complement. To be specific, we implement the integration on the feature level. In the raw image encoder, we enhance its intermediate feature with the current-step encoding features. Each scale of the conditional feature map $m_{I}^{k}$ is fused with the $x_{t}$ encoding features $m_{x}^{k}$ with the same shape, $k$ is the index of layer. The fusion is implemented by an attentive-like mechanism $\mathcal{A}$. In particular, two feature maps are first applied layer normalization and multiply together to get an affinity map. Then we multiply the affinity map with the condition encoding features to enhance the attentive region, which is:
\begin{equation}
    \mathcal{A}(m_{I}^{k}, m_{x}^{k}) = (LN(m_{I}^{k}) \otimes LN(m_{x}^{k})) \otimes m_{I}^{k},
\end{equation}
where $\otimes$ implies element-wise multiplication, $LN$ denotes layer normalization. The operation is applied on the middle two stages, where each is the convolutional stage implemented following ResNet34. 
Such a strategy helps MedSegDiff dynamically localize and calibrate the segmentation. Although effective the strategy it is, another specific problem is that integrating $x_{t}$ embedding will induce extra high-frequency noise. To address this problem, we propose FF-Parser to constrain the high-frequency components in the features.

\subsection{FF-Parser}
We connect FF-parser in the path ways of the feature integration. The function of it is to constrain the noise-related components in the $x_{t}$ features. Our main idea is to learn a parameterized attentive (weight) map applying on the Fourier-space features. Given a decoder feature map $m \in \mathbb{R}^{H \times W \times C}$, we first perform 2D FFT(fast fourier transform) along the spatial dimensions, which we can represented as:
\begin{equation}
    M = \mathcal{F}[m] \in \mathbb{C}^{H \times W \times C},
\end{equation}
where $\mathcal{F}[\cdot]$ denotes the 2D FFT. We then modulate the spectrum of $m$ by multiplying a parameterized attentive map $A \in \mathbb{C}^{H \times W \times C}$ to $M$:
\begin{equation}
    M' = A \otimes M,
\end{equation}
where $\otimes$ denotes the element-wise product. 
Finally, we reverse $M'$ back to the spatial domain by adopting inverse FFT:
\begin{equation}
    m' = \mathcal{F}^{-1}[M'].
\end{equation}
FF-Parser can be regarded as a learnable version of frequency filters which are wildly applied in the digital image processing \cite{pitas2000digital}. Different from the spacial attention, it globally adjusts the components of the specific frequencies. Thus it can be learn to constrain the high-frequency component for the adaptive integration.

\begin{figure}
    \centering
    \includegraphics[width = 0.8 \linewidth]{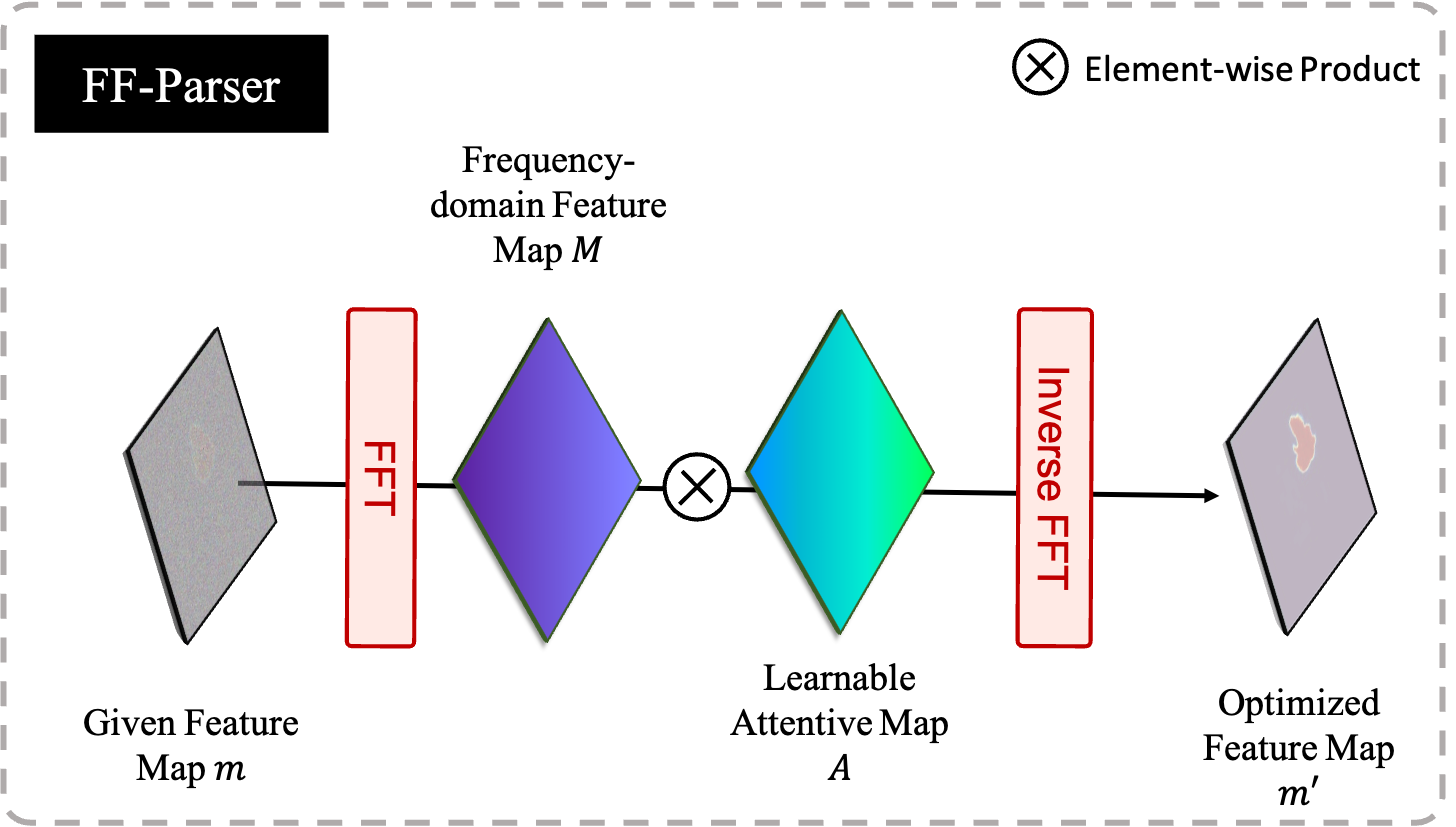}
    \caption{An illustration of FF-Parser. FFT denotes Fast Fourier Transform.}
    \label{fig:parser}
\end{figure}

\subsection{Training and Architecture}\label{AA}
MedSegDiff is trained following the standard process of DPM \cite{ho2020denoising}. Specifically, the loss can be represented as:
\begin{equation}
    \mathcal{L} = E_{x_{0}, \epsilon, t}[||\epsilon - \epsilon_{\theta}(\sqrt{\hat{a}_{t}} x_{0} + \sqrt{1 - \hat{a}_{t}} \epsilon, I_{i}, t)||^{2}].
\end{equation}
In each of the iteration, a random couple of raw image $I_{i}$ and segmentation label $S_{i}$ will be sampled for the training. The iteration number is sampled from a uniform distribution and $\epsilon$ from a Gaussian distribution. 

The main architecture of MedSegDiff is a modified ResUNet\cite{yu2019robust}, which we implement it with a ResNet encoder following a UNet decoder. The detailed network setting is following \cite{nichol2021improved}. $I$ and $x_{t}$ are encoded with two individual encoders. The encoder is consisted of three convolution stages. Each stage contains several residual blocks. The number of residual blocks in each stage is following that of ResNet34. Each residual block is composed of two convolutional blocks, each one consists of group-norm and SiLU\cite{elfwing2018sigmoid} active layer and a convolutional layer. The residual block receives the time embedding through a linear layer, SiLU activation, and another linear layer. The result is then added to the output of the first convolutional block. The obtained $E^{I}$ and $E^{x_{t}}$ are added together and sent to the last encoding stage. A standard convolutional decoder is connected to predict the final result.

\section{Experiments}
\subsection{Dataset}
We conduct the experiments on three different medical tasks with different image modalities, which are optic-cup segmentation from fundus images, brain tumor segmentation from MRI images, and thyroid nodule segmentation from ultrasound images. The experiments of glaucoma, thyroid cancer and melanoma diagnosis are conducted on REFUGE-2 dataset \cite{fang2022REFUGE2}, BraTs-2021 dataset \cite{baid2021rsna} and DDTI dataset \cite{pedraza2015open}, which contain 1200, 2000, 8046 samples, respectively. The datasets are publicly available with both segmentation and diagnosis labels. Train/validation/test sets are split following the default settings of the dataset.

\subsection{Implementation Details}
We experiment with huge, large, basic, and small variants of our model, \textit{MedSegDiff++}, \textit{MedSegDiff-L}, \textit{MedSegDiff-B}, and \textit{MedSegDiff-S}, respectively. \\
In \textit{MedSegDiff-S}, \textit{MedSegDiff-B} \textit{MedSegDiff-L}, \textit{MedSegDiff++}, we use UNet with $4$x, $5$x, $6$x, $6$x downsamples respectively. In the experiments, we employ 100 diffusion steps for the inference, which is much smaller than most of the previous studies\cite{ho2020denoising,nichol2021improved}. All the experiments are implemented with the PyTorch platform and trained/tested on 4 Tesla P40 GPU with 24GB of memory except \textit{MedSegDiff++} and \textit{MedSegDiff-L}. All images are uniformly resized to the dimension of 256$\times$256 pixels. The networks are trained in an end-to-end manner using AdamW\cite{loshchilov2017decoupled} optimizer. \textit{MedSegDiff-B} and \textit{MedSegDiff-S} are trained with 32 batch size, \textit{MedSegDiff-L} and \textit{MedSegDiff++} are trained with 64 batch size. The learning rate is initially set to 1 $\times 10^{-4}$. All models are set 25 times of ensemble in the inference. We use STAPLE\cite{warfield2004simultaneous} algorithm to fuse the different samples. The diffusion based competitor EnsemDiff\cite{wolleb2021diffusion} is reproduced with the same setting for the fair comparison. 

\subsection{Main Results}
We compare with SOTA segmentation methods proposed for the three specific tasks and general medical image segmentation methods. The main results are shown in Table \ref{tab:main}. In the table, ResUnet\cite{yu2019robust} and BEAL\cite{wang2019boundary} are proposed for optic disc/cup segmentation, TransBTS\cite{wang2021transbts} and EnsemDiff\cite{wolleb2021diffusion} are proposed for the brain tumor segmentation, MTSeg\cite{gong2021multi} and UltraUNet\cite{chu2021ultrasonic} are proposed for the Thyroid Nodule segmentation, CENet\cite{gu2019net}, MRNet\cite{ji2021learning}, SegNet\cite{badrinarayanan2017segnet}, nnUNet\cite{isensee2021nnu} and TransUNet\cite{chen2021transunet} are proposed for the general medical image segmentation. We evaluate the segmentation performance by Dice score and IoU.

In Table \ref{tab:main}, we compare with the methods implemented with various network architectures, including CNN (ResUNet, BEAL, nnUNet, SegNet), vision transformer (TransBTS, TransUNet) and DPM (EnsemDiff). We can see the advanced network architectures commonly gain better results. For example, in optic-cup segmentation, ViT-based general segmentation method: TransUNet is even better than the CNN-based task toward method: BEAL. On brain tumor segmentation, recently proposed DPM-based segmentation method EnsemDiff outperforms all those previous ViT-based competitors, i.e., TransBTS and TransUNet. MedSegDiff not only adopts the recent successful DPM, but also designs an appropriate strategy over it specifically towards the general medical image segmentation task. We can see MedSegDiff outperforms all the other methods on three different tasks, which shows the generalization toward different medical segmentation tasks and different image modalities. Comparing against DPM-based model proposed specifically for the brain tumor segmentation, i.e., EnsemDiff, it improves 2.3\% on Dice and 2.4\% on IoU, which indicates the effectiveness of our unique techniques, i.e, dynamic conditioning and FF-Parser. 

\begin{figure}
    \centering
    \includegraphics[width=\linewidth]{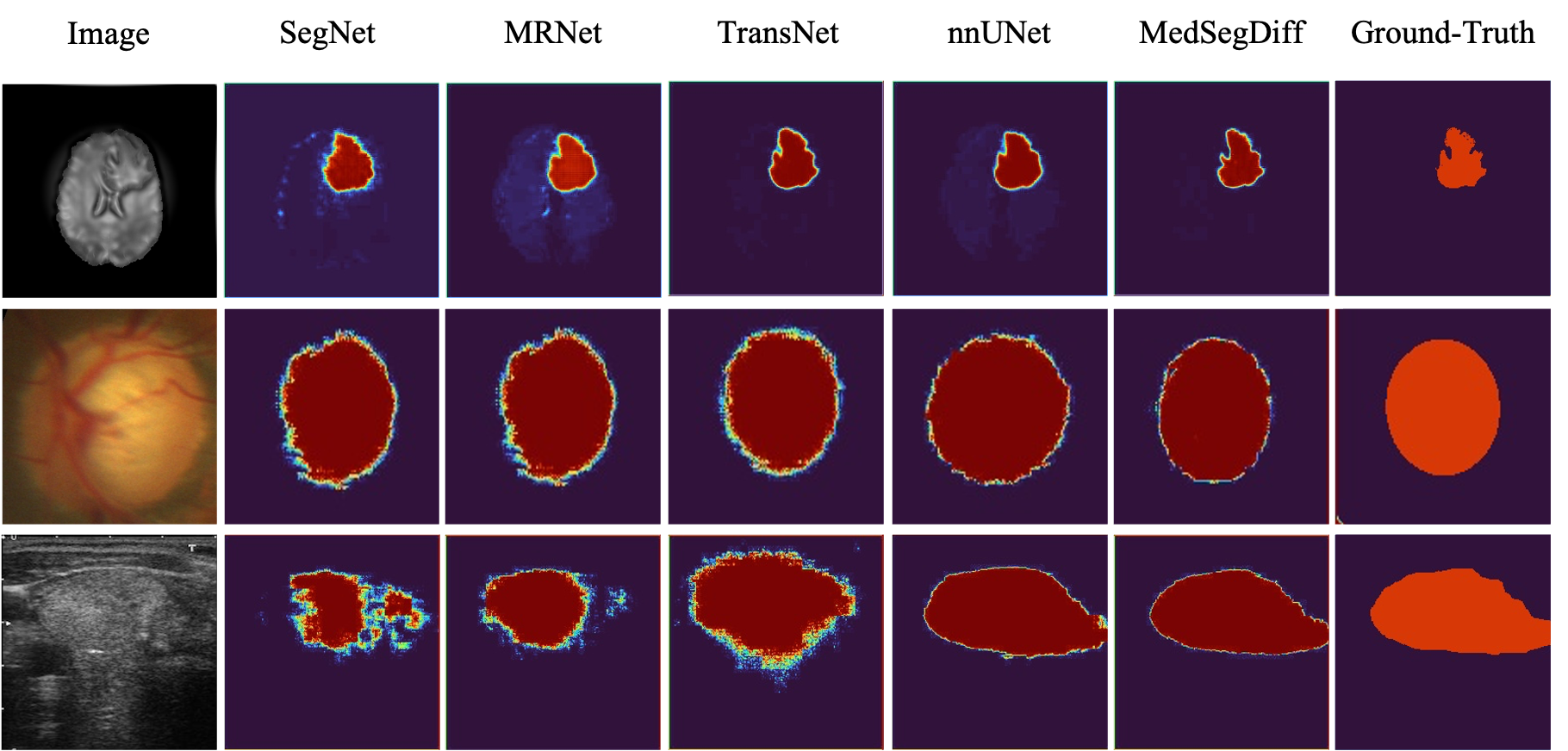}
    \caption{The visual comparison of Top-4 general medical image segmentation methods in Table \ref{tab:main}. From top to down are brain-tumor segmentation, optic-cup segmentation and thyroid nodule segmentation, respectively.}
    \label{fig:vis}
\end{figure}

Figure \ref{fig:vis} shows several typical examples generated by our MedSegDiff and other SOTA methods. It can be seen the target lesions/tissues are all ambiguous on the images so that they are hard to be recognized by human eyes. Comparing with these computer-aided methods, it is obvious that the segmentation maps generated by the proposed method are more accurate than the other methods, especially for the ambiguous regions. To be benefited from DPM together with the proposed dynamic conditioning and FF-Parser, it can better localize and calibrate the segmentation on the low-contrast or ambiguous images.

\begin{table}[h]
\centering
\caption{The comparison of MedSegDiff with SOTA segmentation methods. Best results are denoted as \textbf{bold}. The grey background denotes the methods are proposed for that/these particular tasks.}
\resizebox{0.6\columnwidth}{!}{%
\begin{tabular}{c|cc|cc|cc}
\hline
{\color[HTML]{333333} }          & \multicolumn{2}{c|}{{\color[HTML]{333333} Optic-Cup}}                                                     & \multicolumn{2}{c|}{{\color[HTML]{333333} Brain-Turmor}}                                                  & \multicolumn{2}{c}{{\color[HTML]{333333} Thyroid Nodule}}                                                 \\ \hline
{\color[HTML]{333333} }          & {\color[HTML]{333333} Dice}                         & {\color[HTML]{333333} IoU}                          & {\color[HTML]{333333} Dice}                         & {\color[HTML]{333333} IoU}                          & {\color[HTML]{333333} Dice}                         & {\color[HTML]{333333} IoU}                          \\ \hline
{\color[HTML]{333333} ResUnet}   & \cellcolor[HTML]{EFEFEF}{\color[HTML]{333333} 80.1} & \cellcolor[HTML]{EFEFEF}{\color[HTML]{333333} 72.3} & {\color[HTML]{333333} -}                            & {\color[HTML]{333333} -}                            & {\color[HTML]{333333} -}                            & {\color[HTML]{333333} -}                            \\
{\color[HTML]{333333} BEAL}      & \cellcolor[HTML]{EFEFEF}{\color[HTML]{333333} 83.5} & \cellcolor[HTML]{EFEFEF}{\color[HTML]{333333} 74.1} & {\color[HTML]{333333} -}                            & {\color[HTML]{333333} -}                            & {\color[HTML]{333333} -}                            & {\color[HTML]{333333} -}                            \\ \hline
{\color[HTML]{333333} TransBTS}  & {\color[HTML]{333333} -}                            & {\color[HTML]{333333} -}                            & \cellcolor[HTML]{EFEFEF}{\color[HTML]{333333} 87.6} & \cellcolor[HTML]{EFEFEF}{\color[HTML]{333333} 78.3} & {\color[HTML]{333333} -}                            & {\color[HTML]{333333} -}                            \\
{\color[HTML]{333333} EnsemDiff} & {\color[HTML]{333333} -}                            & {\color[HTML]{333333} -}                            & \cellcolor[HTML]{EFEFEF}{\color[HTML]{333333} 88.7} & \cellcolor[HTML]{EFEFEF}{\color[HTML]{333333} 80.9} & {\color[HTML]{333333} -}                            & {\color[HTML]{333333} -}                            \\ \hline
{\color[HTML]{333333} MTSeg}     & {\color[HTML]{333333} -}                            & {\color[HTML]{333333} -}                            & {\color[HTML]{333333} -}                            & {\color[HTML]{333333} -}                            & \cellcolor[HTML]{EFEFEF}{\color[HTML]{333333} 82.3} & \cellcolor[HTML]{EFEFEF}{\color[HTML]{333333} 75.2} \\
{\color[HTML]{333333} UltraUNet} & {\color[HTML]{333333} -}                            & {\color[HTML]{333333} -}                            & {\color[HTML]{333333} -}                            & {\color[HTML]{333333} -}                            & \cellcolor[HTML]{EFEFEF}{\color[HTML]{333333} 84.5} & \cellcolor[HTML]{EFEFEF}{\color[HTML]{333333} 76.2} \\ \hline
\rowcolor[HTML]{EFEFEF} 
{\color[HTML]{333333} CENet}     & {\color[HTML]{333333} 78.6}                         & {\color[HTML]{333333} 69,4}                         & {\color[HTML]{333333} 76.2}                         & {\color[HTML]{333333} 68.9}                         & {\color[HTML]{333333} 78.9}                         & {\color[HTML]{333333} 71.2}                         \\
\rowcolor[HTML]{EFEFEF} 
{\color[HTML]{333333} MRNet}     & {\color[HTML]{333333} 84.2}                         & {\color[HTML]{333333} 75.1}                         & {\color[HTML]{333333} 83.4}                         & {\color[HTML]{333333} 75.6}                         & {\color[HTML]{333333} 80.4}                         & {\color[HTML]{333333} 73.4}                         \\
\rowcolor[HTML]{EFEFEF} 
{\color[HTML]{333333} SegNet}    & {\color[HTML]{333333} 80.4}                         & {\color[HTML]{333333} 70.7}                         & {\color[HTML]{333333} 80.2}                         & {\color[HTML]{333333} 72.9}                         & {\color[HTML]{333333} 81.7}                         & {\color[HTML]{333333} 74.5}                         \\
\rowcolor[HTML]{EFEFEF} 
nnUNet                           & 84.9                                                & 75.1                                                & 88.2                                                & 80.4                                                & 84.2                                                & 76.2                                                \\
\rowcolor[HTML]{EFEFEF} 
TransUNet                        & 85.6                                                & 75.9                                                & 86.6                                                & 79.0                                                & {\color[HTML]{333333} 83.5}                         & 75.1                                                \\ \hline
\rowcolor[HTML]{EFEFEF} 
MedSegDiff-S                         & 81.2                                       & 71.7                                       & 82.3                                       & 73.6                                       & 80.8                                       & 73.7                                       \\
\rowcolor[HTML]{EFEFEF} 
MedSegDiff-B                        & 85.9                                       & 76.2                                       & 88.9                                       & 81.2                                       & 84.8                                       & 76.4                                       \\
\rowcolor[HTML]{EFEFEF} 
MedSegDiff-L                       & 86.9                                       & 78.5                                       & 89.9                                       & 82.3                                       & 86.1                                       & 79.6                                       \\
\rowcolor[HTML]{EFEFEF} 
MedSegDiff++                       & \textbf{87.5}                                       & \textbf{79.1}                                       & \textbf{90.5}                                       & \textbf{82.8}                                       & \textbf{86.6}                                       & \textbf{80.2}                                       \\ \hline
\end{tabular}%
}
\label{tab:main}
\end{table}

\subsection{Ablation Study}
We do comprehensive ablation study to verify the effectiveness of the proposed dynamic conditioning and FF-Parser. The results are shown in Table \ref{tab:ab}, where Dy-Cond denotes dynamic conditioning. We evaluate the performance by Dice score(\%) on all three tasks. From the table, we can see Dy-Cond gains considerable improvements over vanilla DPM. On the case which the region localization is important, i.e., optic-cup segmentation, it improves 2.1\%. On the cases which the images are low-contrast, like brain tumor and thyroid nodule segmentation, it improves 1.6\% and 1.8\% respectively. It shows Dy-Cond is a generally effective strategy on DPM for both of the cases. FF-Parser which established over Dy-Cond mitigates the high-frequency noises thus further optimize the segmentation results. It helps MedSegDiff further improve near 1\% performance and achieve the best on all three tasks. 
\begin{table}[h]
\centering
\caption{An ablation study on dynamic condition encoding and FF-Parser. Dice score(\%) is used as the metric.}
\begin{tabular}{cc|ccc}
\hline
Dy-Cond & FF-Parser & OpticCup & BrainTumor & ThyroidNodule \\ \hline
        &           & 84.6     & 88.2       & 84.1          \\
\checkmark       &           & 86.7     & 89.8       & 85.9          \\
\checkmark       & \checkmark         & \textbf{87.5} & \textbf{90.5} & \textbf{86.6} \\ \hline
\end{tabular}\label{tab:ab}
\end{table}

\section{Conclusion}
In this paper, we provided a scheme for DPM-based general medical image segmentation, named MedSegDiff. We propose two novel techniques to promise the performance of it, i.e., the dynamic conditional encoding and FF-Parser. The comparison experiments are conducted on three medical image segmentation tasks with different image modalities, which shows our model outperforms previous SOTA. As the first DPM application in general medical image segmentation, we believe MedSegDiff will serve as an essential benchmark for future research.
\clearpage


\bibliographystyle{splncs04}
\bibliography{egbib}






\end{document}